\documentclass[nohyperref]{article}

\usepackage{microtype}
\usepackage{graphicx}
\usepackage{subfigure}
\usepackage{booktabs} 

\usepackage[backref=page]{hyperref}

\usepackage[accepted]{pre-training}

\usepackage{amsmath}
\usepackage{amssymb}
\usepackage{mathtools}
\usepackage{amsthm}

\usepackage{arydshln} 
\usepackage{multirow} 
\usepackage{wrapfig} 
\usepackage{tabularx} 
\usepackage{booktabs}
\usepackage{float}

\usepackage[capitalize,noabbrev]{cleveref}

\theoremstyle{plain}

\theoremstyle{definition}

\theoremstyle{remark}

\usepackage[textsize=tiny]{todonotes}

\icmltitlerunning{Hyper-Representations for Pre-Training and Transfer Learning}

\begin{document}

\twocolumn[
\icmltitle{Hyper-Representations for Pre-Training and Transfer Learning}



\icmlsetsymbol{equal}{*}

\begin{icmlauthorlist}
\icmlauthor{Konstantin Sch\"urholt}{xxx}
\icmlauthor{Boris Knyazev}{yyy}
\icmlauthor{Xavier Giró-i-Nieto}{zzz}
\icmlauthor{Damian Borth}{xxx}
\end{icmlauthorlist}

\icmlaffiliation{xxx}{AIML Lab, School of Computer Science, University of St. Gallen, St. Gallen, Switzerland}
\icmlaffiliation{yyy}{Samsung - SAIT AI Lab, Montreal, Canada}
\icmlaffiliation{zzz}{Image Processing Group, Universitat Politècnica de Catalunya, Barcelona, Spain}

\icmlcorrespondingauthor{Konstantin Sch\"urholt}{konstantin.schuerholt@unisg.ch}

\icmlkeywords{Machine Learning, ICML, Hyper-Representations, Model Zoo, Representation Learning, Knowledge Transfer}

\vskip 0.3in
]



\printAffiliationsAndNotice{}  

\begin{abstract}
Learning representations of neural network weights given a model zoo is an emerging and challenging area with many potential applications from model inspection, to  neural architecture search or knowledge distillation.
Recently, an autoencoder trained on a model zoo was able to learn a \textit{hyper-representation}, which captures intrinsic and extrinsic properties of the models in the zoo.
In this work, we extend hyper-representations for generative use to sample new model weights as pre-training. 
We propose layer-wise loss normalization which we demonstrate is key to generate high-performing models and a sampling method based on the empirical density of hyper-representations.
The models generated using our methods are diverse, performant and capable to outperform conventional baselines for transfer learning.
Our results indicate the potential of knowledge aggregation from model zoos to new models via hyper-representations thereby paving the avenue for novel research directions.\looseness-1
\end{abstract}

\begin{figure*}[t]
\begin{minipage}[t]{0.98\textwidth}
\begin{center}
\includegraphics[trim=0in 0in 0in 0in, clip, width=1.0\linewidth]{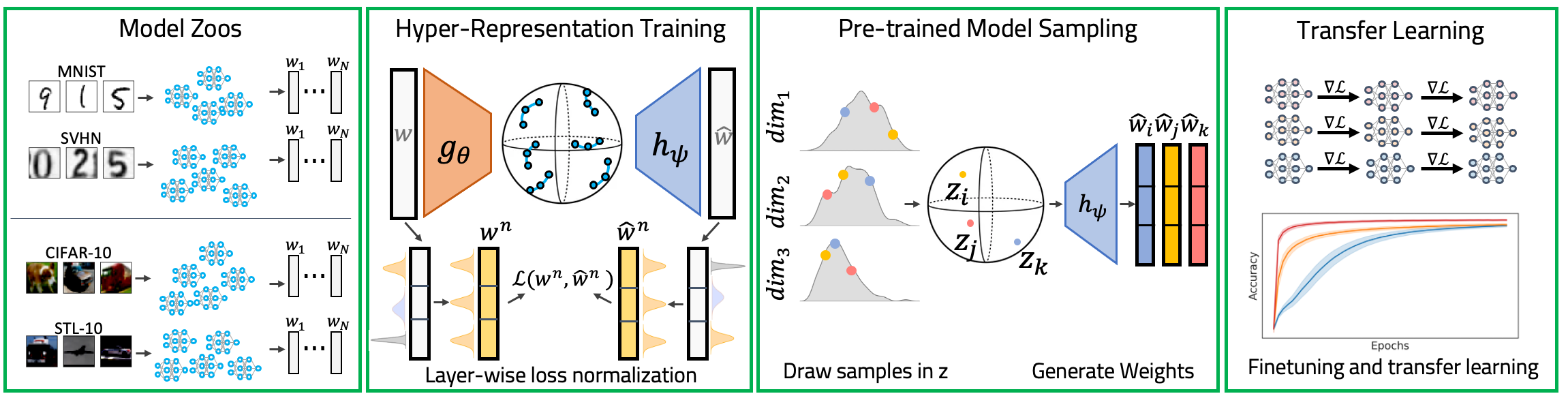}
\vskip -0.1in
\caption{\small \textbf{Outline of our approach}: Model zoos are trained on image classification tasks. Hyper-representations are trained with self-supervised learning on the weights of the model zoos using layer-wise loss normalization in the reconstruction loss. 
We sample new embeddings in hyper-representation space and decode them to neural network weights. 
Generated models are diverse and able to be used as pre-trained models for subsequent transfer learning on new datasets.
}
\label{fig:scheme}    
\end{center}
\end{minipage}
\end{figure*}

%
%
\section{Introduction}
Pre-training is a key element to transfer learning. What if we could encode the knowledge of an 
entire population of neural networks for transfer learning to a new domain or dataset?
Recent work~\citep{unterthinerPredictingNeuralNetwork2020,schurholtSelfSupervisedRepresentationLearning2021,martin2021predicting} demonstrated the capability to predict properties of individual neural networks captured by 
such population of neural networks (often also referred to as \textit{model zoo}).

In particular,~\citep{schurholtSelfSupervisedRepresentationLearning2021} proposed to learn a lower-dimensional
representation of a model zoo able to capture the underlying manifold of all neural network models populating 
the model zoo. In their work, they trained so called \textit{hyper-presentations} with a transformer autoencoder directly 
from the weights of a zoo and used them to predict several model properties such as accuracy, hyperparameters or
architecture configurations of individual neural networks.

Instead of using hyper-presentations for model properties prediction only, in this work, we propose to exploit 
the knowledge encoded in hyper-representations as pre-training for subsequent transfer learning. 
We train hyper-representations of model zoos and use them in combination with the decoder as a generative model
i.e., to sample model weights as pre-trained models in a single forward pass through the decoder. To that end, 
we introduce \textit{layer-wise loss normalization} improving the quality of decoded neural network weights 
significantly and demonstrate that conditioning the proposed sampling methods on particular properties of the 
topology of the hyper-representation makes a difference for transfer learning. Our results show that the proposed 
approach is potent enough to be used as another form of pre-training able to out-perform conventional baselines
in transfer learning.

Previous work on generating model weights proposed (Graph) HyperNetworks~\citep{haHyperNetworks2016,zhangGraphHyperNetworksNeural2020,knyazevParameterPredictionUnseen2021}, Bayesian HyperNetworks~\citep{deutschGeneratingNeuralNetworks2018}, HyperGANs~\citep{ratzlaffHyperGANGenerativeModel2019} and HyperTransformers~\citep{zhmoginovHyperTransformerModelGeneration2022} for neural architecture search, model compression, ensembling, transfer- or meta-learning.
These methods learn representations from the images and labels of the target domain. In contrast, our approach only uses model weights and does not need access to the underlying data samples and labels rendering it more compact to the original data. In addition to the ability to generate novel and diverse model weights, compared to previous works our approach (a) can generate novel weights conditionally on model zoos for unseen datasets and (b) can be conditioned on the latent factors of the underlying hyper-representation. Notably, both (a) and (b) can be done without the need to retrain hyper-representations.\looseness-1

The results suggest our approach (Figure \ref{fig:scheme}) to be a promising step towards the use of 
hyper-representation as generative models able to encapsulate knowledge of model zoos for transfer learning.

%
%
\section{Hyper-Representation Training}
\label{sec:training}

The first stage of our method that corresponds to learning a hyper-representation of a population of neural networks, called a \textit{model zoo}~\citep{schurholtSelfSupervisedRepresentationLearning2021}. In this context, a model zoo consists of models of the same architecture trained on the same task such as CIFAR-10 image classification~\citep{krizhevskyLearningMultipleLayers2009}.
Specifically, a hyper-representation is learned using an autoencoder $\hat{\mathbf{w}}_i = h(g(\mathbf{w}_i))$ on a zoo of $M$ models $\{ {\bf w}_i\}_1^M$, 
where ${\bf w}_i$ is the flattened vector of dimension $N$ of all the weights of the $i$-th model. 
The encoder $g$ compresses vector $\mathbf{w}_i$ to fixed-size hyper-representation $\mathbf{z}_i=g(\mathbf{w}_i)$ of lower dimension. The decoder $h$ decompresses the hyper-representation to the reconstructed vector $\hat{\mathbf{w}}_i$. Both encoder and decoder are built on a self-attention blocks. The samples from model zoos are understood as sequences of convolutional or fully connected neurons. Each of the these is encoded as a token embedding and concatenated to a sequence. 
The sequence is passed through several layers of multi-head self-attention. Afterwards, a special compression token summarizing the entire sequence is linearly compressed to the bottleneck. The output is fed through a tanh-activation to achieve a bounded latent space $\mathbf{z}_i$ for the hyper-representation. The decoder is symmetric to the encoder, the embeddings are linearly decompressed from hyper-representations $\mathbf{z}_i$ and position encodings added.

Training is done in a multi-task fashion, minimizing the composite loss $\mathcal{L} = \beta \mathcal{L}_{MSE}+(1-\beta)\mathcal{L}_{c}$, where
$\mathcal{L}_{c}$ is a contrastive loss and $\mathcal{L}_{MSE}$ is a weight reconstruction loss (see details in~\citep{schurholtSelfSupervisedRepresentationLearning2021}). 
We can write the latter in a layer-wise way to facilitate our discussion in \S~\ref{sec:lwln}:\looseness-1
\begin{equation}
    \label{eq:baseline_loss}
    \mathcal{L}_{MSE} = \frac{1}{MN}\sum\nolimits_{i=1}^M \sum\nolimits_{l=1}^L || \hat{\mathbf{w}}^{(l)}_i - {\mathbf{w}}^{(l)}_i ||^2_2,
\end{equation}
\noindent where $\hat{\mathbf{w}}^{(l)}_i$, ${\mathbf{w}}^{(l)}_i$ are reconstructed and original weights for the $l$-{th} layer of the $i$-{th} model in the zoo; $L$ is the total number of layers. 
The contrastive loss $\mathcal{L}_{c}$ leverages two types of data augmentation at train time to impose structure on the latent space: permutation exploiting inherent symmetries of the weight space and random erasing.

\section{Methods}
\label{sec:methods}
\subsection{Layer-Wise Loss Normalization}\label{sec:lwln}
We observed that hyper-representations as proposed by~\citep{schurholtSelfSupervisedRepresentationLearning2021} decode to dysfunctional models, with performance around random guessing. 
To alleviate that, we propose a novel layer-wise loss normalization (LWLN), which we motivate and detail in the following.
\begin{figure}[h!]
\begin{minipage}[t]{0.5\textwidth}
\begin{center}
\vspace{-4mm} 
\includegraphics[trim=0mm 0mm 0mm 0mm, clip, width=1.0\linewidth]{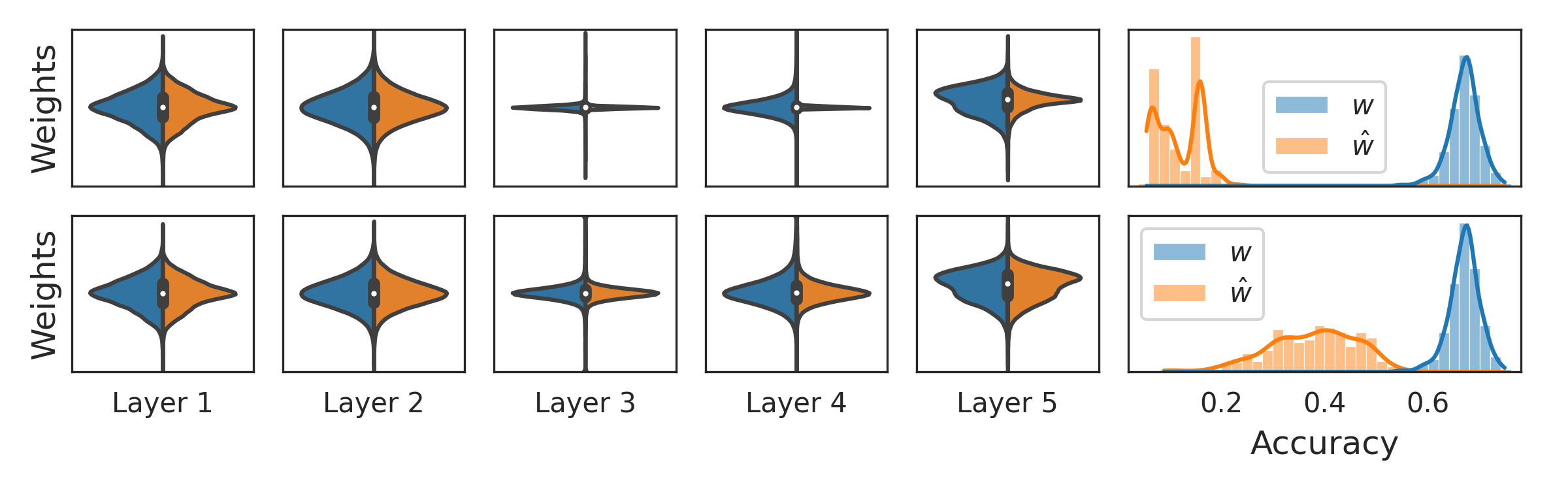}
\vspace{-10mm} 
\caption{
\small Comparison of the distributions of SVHN zoo weights $\mathbf{w}$ (blue) and reconstructed weights $\hat{\mathbf{w}}$ (orange) as well as their accuracy on the SVHN test set. 
\textbf{Top:} Baseline hyper-representation as proposed by~\citep{schurholtSelfSupervisedRepresentationLearning2021}, the weights of layers 3, 4 collapse to the mean. These layers form a weak link in reconstructed models. The accuracy of reconstructed models drops to random guessing.
\textbf{Bottom:} Hyper-representation trained with layer-wise loss normalization (LWLN). The normalized distributions are balanced, all layers are evenly reconstructed, and the accuracy of reconstructed models is significantly improved.
}
\vspace{-3mm} 
\label{fig:layer_norm_eval}    
\end{center}
\end{minipage}
\end{figure}

Due to the MSE training loss, the reconstruction error can generally be expected to be uniformly distributed over all weights and layers of the weight vector $\mathbf{w}$.
However, the weight magnitudes of many of our zoos are unevenly distributed across different layers. 
In these zoos, the even distribution of reconstruction errors lead to undesired effects.
Layers with broader distributions and large weights are reconstructed well, while layers with narrow distributions and small weights are disregarded. The latter can become a weak link in the reconstructed models, causing performance to drop significantly down to random guessing. 
The top row of Figure \ref{fig:layer_norm_eval} shows an example of a baseline hyper-representation learned on the zoo of SVHN models~\citep{netzerReadingDigitsNatural2011}.
Common initialization schemes~\citep{heDelvingDeepRectifiers2015,glorotUnderstandingDifficultyTraining2010} produce distributions with different scaling factors per layer, so the issue is not an artifact of the zoos, but can exist in real world model populations. Similarly, recent work on generating models normalizes weights to boost performance~\citep{knyazevParameterPredictionUnseen2021}.
In order to achieve equally accurate reconstruction across the layers, we introduce a layer-wise loss normalization (LWLN) with the mean $\mu_l$ and standard deviation $\sigma_l$ of all weights in layer $l$ estimated over the train split of the zoo:\looseness-1
\begin{equation}
    \label{eq:lwln_loss}
    \mathcal{L}_{\bar{MSE}} = \frac{1}{MN}\sum_{i=1}^M \sum_{l =1}^{L}\frac{\|\hat{\mathbf{w}}_i^{(l)}-\mathbf{w}_i^{(l)} \|_2^2}{\sigma_l^2}. 
\end{equation}

\subsection{Sampling from Hyper-Representations}
\label{sec:sampling}
We introduce methods to draw diverse and high-quality samples $\mathbf{z}^* \sim p(\mathbf{z})$ from the learned hyper-representation space to generate model weights $\mathbf{w}^* = h(\mathbf{z}^*)$. 
%
Assuming the smoothness and robustness of the learned hyper-representation space, we follow ~\citep{liu2019acceleration,guo2019auto,ghoshVariationalDeterministicAutoencoders2020} in estimating the density to draw samples from a regularized autoencoder. 
We propose to draw samples from the density of the train set in representation space. We use the embeddings of the train set as anchor samples $\{\mathbf{z}_i\}$ to model the density. 
The dimensionality $D$ of hyper-representations $\mathbf{z}$ in~\citep{schurholtSelfSupervisedRepresentationLearning2021}, as well as in our work, is relatively high due to the challenge of compressing weights $\mathbf{w}$. 
We make a conditional independence assumption to facilitate sampling: $p(\mathbf{z}^{(j)}|\mathbf{z}^{(k)}, \mathbf{w}) = p(\mathbf{z}^{(j)}| \mathbf{w})$, where $\mathbf{z}^{(j)}$ is the $j$-th dimensionality of the embedding $\mathbf{z}$.
To model the distribution of each $j$-th dimensionality, we choose kernel density estimation (KDE), as it is a powerful yet simple, non-parametric and deterministic method with a single hyperparameter.
We fit a KDE to the $M$ anchor samples $\{\mathbf{z}_{i}^{(j)}\}_{i=1}^M$ of each dimension $j$, and draw samples $z^{(j)}$ from that distribution: $ z^{(j)} \sim p(\mathbf{z}^{(j)}) = \frac{1}{M h} \sum\nolimits_{i=1}^{M} K(\frac{\mathbf{z}^{(j)} - \mathbf{z}^{(j)}_i}{h})
$,
\noindent where $K(x)=(2\pi)^{-1/2} \exp{(-\frac{x^2}{2})}  $
is the Gaussian kernel and $h$ is a bandwidth hyperparameter.
The samples of each dimension $z^{(j)}$ are concatenated to form samples $\mathbf{z}^* = [z^{(1)}, z^{(2)}, \cdots, z^{(D)}]$. This method is denoted as $S_{\text{KDE}}$.
We observe that many anchor samples from $\{\mathbf{z}_i\}$ correspond to the models with relatively poor accuracy (Figure \ref{fig:layer_norm_eval}). To improve the quality of sampled weights, we consider the variants of $S_{\text{KDE}}$ using only those embeddings of training samples corresponding to the top 30\% performing models. We denote that sampling method as $S_{\text{KDE30}}$.\looseness-1

%
\section{Experiments}
\label{sec:experiments}

\subsection{Experimental Setup}

We train and evaluate our approaches on four image classification datasets: MNIST~\citep{lecunGradientbasedLearningApplied1998}, SVHN~\citep{netzerReadingDigitsNatural2011}, 
CIFAR-10~\citep{krizhevskyLearningMultipleLayers2009} and STL-10~\citep{coatesAnalysisSingleLayerNetworks2011}. For each dataset, there is a model zoo that we use to train an autoencoder.

\textbf{Model zoos:} 
In each image dataset,  a zoo contains $M=1000$ convolutional networks of the same architecture with three convolutional layers and two fully-connected layers ($L=5$). 
Varying only in the random seeds, all models of the zoo are trained for 50 epochs with the same hyperparameters following~\citep{schurholtSelfSupervisedRepresentationLearning2021}. 
To integrate higher diversity in the zoo, initial weights are uniformly sampled from a wider range of values rather than using well-tuned initializations of~\citep{glorotUnderstandingDifficultyTraining2010,heDelvingDeepRectifiers2015}.
Each zoo is split in the train (70\%), validation (15\%) and test (15\%) splits. 
To incorporate the learning dynamics, we train autoencoders on the models trained for
21-25 epochs following~\citep{schurholtSelfSupervisedRepresentationLearning2021}. Here the models have already achieved high performance, but have not fully converged. The development in the remaining epochs of each model is treated as hold-out data to compare against. 
We use the MNIST and SVHN zoos from~\citep{schurholtSelfSupervisedRepresentationLearning2021} and based on them create the CIFAR-10 and STL-10 zoos. Details on the zoos can be found in Appendix~\ref{app:zoos}.\looseness-1

\textbf{Self-Supervised Hyper-Representation Training:} We train separate hyper-representations on each of the model zoos and use the checkpoint with lowest reconstruction error on the validation set. 
Using the proposed sampling methods (\S~\ref{sec:sampling}), we generate new embeddings and decode them to weights. We evaluate sampled populations as initializations (epoch 0) and by fine-tuning for up to 25 epochs.
We distinguish between in-dataset and transfer-learning. 
For in-dataset, the same image dataset is used for training and evaluating our hyper-representations and baselines.
For transfer-learning, pre-trained models $B_F$ and the hyper-representation model are trained on a source dataset. Subsequently, the pre-trained models $B_F$ and the samples $S_{\text{KDE}}$ are fine-tuned on the target domain. The baseline ($B_T$) contains models trained from scratch on the target domain.

%
\textbf{Baselines:}
As the first baseline, we consider the autoencoder of~\citep{schurholtSelfSupervisedRepresentationLearning2021}, which is same as ours but without the proposed layer-wise loss-normalization (LWLN, \S~\ref{sec:lwln}). We combine this autoencoder with the $S_\text{KDE30}$ sampling method and, hence, denote it as $B_{\text{KDE}30}$.
We consider two other baselines based on training models with stochastic gradient descent (SGD): training from scratch on the target classification task $B_T$, and training on a source followed by fine-tuning on the target task $B_F$. The latter remains one of the strongest transfer learning baselines~\citep{chen2019closer,dhillon2019baseline,kolesnikov2020big}.

%
\subsection{Results}
\label{sec:results}
\begin{table}[]
\caption{
Mean and std accuracy (\%) of sampled populations with LWLN ($S_{\text{KDE}30}$) and without ($B_{\text{KDE}30}$) compared to models trained from scratch $B_T$. 
}\label{tab:initialization}
\vspace{2mm}
\scriptsize
\setlength{\tabcolsep}{6pt}
\begin{tabularx}{\linewidth}{rccccc}
\toprule
\multicolumn{1}{r}{\textbf{Method}} & \textbf{Ep.} & \multicolumn{1}{c}{\textbf{MNIST}} & \multicolumn{1}{c}{\textbf{SVHN}} &
\multicolumn{1}{c}{\textbf{CIFAR-10}} & \multicolumn{1}{c}{\textbf{STL-10}} \\
\midrule
$B_{T}$                & 0 &  \multicolumn{4}{c}{$\approx$10\% (random guessing)}\\
$B_{\text{KDE}30}$   & 0     & {63.2 \textit{(7.2)}}   & 10.1 \textit{(3.2)}     & 15.5 \textit{(3.4)} & 12.7  \textit{(3.4)}  \\
$S_{\text{KDE}}$     & 0     & 66.4 \textit{(7.3)}   & 46.7 \textit{(8.3)} & 24.8 \textit{(5.1)} & 18.9 \textit{(2.1)} \\
$S_{\text{KDE}30}$   & 0     & \textbf{68.6 \textit{(6.7)}}   & \textbf{51.5 \textit{(5.9)}}  & \textbf{26.9  \textit{(4.9)}}  & \textbf{19.7 \textit{(2.1)}}  \\
\midrule
$B_{T}$                & 1      & 20.6 \textit{(1.6)}  & 19.4 \textit{(0.6)}  & 27.5 \textit{(2.1)} & 15.4 \textit{(1.8)}  \\
$B_{\text{KDE}30}$   & 1     & {83.2 \textit{(1.2)}}   & 67.4 \textit{(2.0)} & 39.7 \textit{(0.6)} & \textbf{26.4 \textit{(1.6)}} \\
$S_{\text{KDE}}$     & 1     & 80.4 \textit{(3.2)}   & 66.2 \textit{(8.2)} & 43.3 \textit{(1.3)} & 24.1 \textit{(2.1)} \\
$S_{\text{KDE}30}$   & 1     & \textbf{83.7 \textit{(1.3)}}   & \textbf{69.9 \textit{(1.6)}}  & \textbf{44.0 \textit{(0.5)}}  &	{25.9 \textit{(1.6)}}  \\
\midrule
$B_{T}$                & 25     & 83.3 \textit{(2.6)} & 66.7 \textit{(8.5)} & 46.1 \textit{(1.3)} & 35.0 \textit{(1.3)}  \\
$B_{\text{KDE}30}$   & 25    & \textbf{93.2 \textit{(0.6)}}   & \textbf{75.4 \textit{(0.9)}}  & {48.1 \textit{(0.6)}}  & \textbf{38.4 \textit{(0.9)}}  \\
$S_{\text{KDE}}$     & 25    & 92.5 \textit{(0.8)}   & 71.8 \textit{(7.7)} & 48.0 \textit{(1.2)} & 37.4 \textit{(1.3)} \\
$S_{\text{KDE}30}$   & 25    & {93.0 \textit{(0.7)}}    & {74.2 \textit{(1.4)}} & \textbf{48.6 \textit{(0.5)}}  & {38.1 \textit{(1.1)}}  \\
\midrule
$B_{T}$                & 50     & 91.1 \textit{(2.6)} & 70.7 \textit{(8.8)} & 48.7 \textit{(1.4)} & 39.0 \textit{(1.0)}  \\
\bottomrule 
\end{tabularx}
\vspace{-6mm}
\end{table} 

\textbf{Evaluation of layer-wise loss normalization:}  We compare $S_{\text{KDE30}}$ that is based on our autoencoder with layer-wise loss normalization (LWLN) to the baseline autoencoder using the same sampling method ($B_{\text{KDE}30}$) without fine-tuning. 
On all datasets except for MNIST, $S_{\text{KDE30}}$ considerably outperform $B_{\text{KDE}30}$ with the latter performing just above 10\% (random guessing), see Table~\ref{tab:initialization} (rows with epoch 0).
We attribute the success of LWLN to two main factors.
First, LWLN prevents the collapse of reconstruction to the mean (compare Figure \ref{fig:layer_norm_eval}-top to Figure \ref{fig:layer_norm_eval}-bottom). Second, by fixing the weak links, the reconstructed models perform significantly better.
We also evaluated input-to-output feature normalization, s.t. encoder and decoder operate on normalized weights, but empirically found it did not work as well as a normalization just for the loss.

\textbf{Sampling for in-dataset fine-tuning:} 
When fine-tuning, our $S_{\text{KDE}30}$ and baseline $B_{\text{KDE}30}$ appear to gradually converge to similar performance (Table \ref{tab:initialization}). While unfortunate, this result aligns well with previous findings that longer training and enough data make initialization less important~\citep{mishkin2015all,he2019rethinking,rasmus2015semi}.
Comparing $S_{\text{KDE}}$ and $S_{\text{KDE}30}$, we observe that conditioning the samples on the better models in the zoo improves the performance.
We also compare $S_{\text{KDE}}$ and $S_{\text{KDE}30}$ to training models from scratch ($B_T$).
On all four datasets, both ours and the baseline hyper-representations outperform $B_T$ when generated weights are fine-tuned for the same number of epochs as $B_T$.
Notably, on MNIST and SVHN generated weights fine-tuned for 25 epochs are even better than $B_T$ run for 50 epochs. The comparison to $B_T$ trained for 50 epochs on the image dataset is interesting, since the hyper-representations were trained on model weights trained for up to 25 epochs, and so their overall training epochs on the image dataset is equal.
On CIFAR-10 and STL-10, all populations are limited by the architecture and saturate below 50 and 40 \% accuracy.
These findings show that the models initialized with generated weights can learn faster and in some cases achieve higher performance in 25 epochs than $B_T$ in 50 epochs. \looseness-1

\paragraph{Sampling for Cross-dataset Initialization}
%
We investigate the effectiveness of our method in a transfer-learning setup across image datasets. 
Here, a zoo is trained on a source dataset, e.g., SVHN. A hyper-representation is trained on that zoo and models are generated from it. These models are transferred to a target dataset, e.g., MNIST.
We report transfer learning results from SVHN to MNIST and from STL-10 to CIFAR-10 as two representative scenarios. Results on all datasets can be found in Appendix \ref{app:results}. 
\begin{table}[ht!]
\centering
\caption{Transfer-learning results (mean and std accuracy in \%). Note that for STL-10 to CIFAR-10 the performance of all methods saturate quickly due to the limited capacity of models in the zoo making further improvements challenging.
}
\label{tab:transfer}
\vspace{2mm}
\scriptsize
\setlength{\tabcolsep}{10pt}
\begin{tabularx}{\linewidth}{rccc}
\toprule
\multicolumn{1}{r}{\textbf{Method}} & \textbf{Ep.} & \multicolumn{1}{c}{\textbf{SVHN to MNIST}} & \multicolumn{1}{c}{\textbf{STL-10 to CIFAR-10}} \\
\midrule
$B_{T}$                & 0 &  \multicolumn{2}{c}{$\approx$10\% (random guessing)}\\
$B_{F}$                & 0 & \textbf{33.4 \textit{(5.4)}}   & \textbf{15.3 (2.3)}          \\
$S_{\text{KDE}30}$     & 0 &  31.8 \textit{(5.6)}    & 14.5 \textit{(1.9)}             \\
\midrule
$B_{T}$                & 1      & 20.6 \textit{(1.6)}        & 27.5 \textit{(2.1)}  \\
$B_{F}$                & 1      & 84.4 \textit{(7.4)}    & 29.4 \textit{(1.9)}  \\
$S_{\text{KDE}30}$     & 1      & \textbf{86.9 \textit{(1.4)}}      & \textbf{29.6 \textit{(2.0)}}  \\
\midrule
$B_{T}$                & 50     & 91.1 \textit{(1.0)}       & 48.7 \textit{(1.4)} \\
$B_{F}$                & 50     & 95.0 \textit{(0.8)}        & \textbf{49.2 \textit{(0.7)}} \\
$S_{\text{KDE}30}$     & 50    & \textbf{95.5 \textit{(0.7)}}       & 48.8 \textit{(0.9)} \\
\bottomrule 
\end{tabularx}
\vspace{-4mm}
\end{table} 
%
In transfer learning from SVHN to MNIST, the sampled populations on average learn faster and achieve significantly higher performance than the $B_T$ baseline and generally compares favorably to $B_F$ 
(Table~\ref{tab:transfer}). 
In the STL-10 to CIFAR-10 experiment, all populations appear to saturate with only small differences in their performances (Table \ref{tab:transfer}).
We found that all datasets are useful sources for all targets (see Appendix \ref{app:results}).
This might be explained by the ability of hyper-representations to capture a generic inductive prior useful across different domains.\looseness-1 

\section{Conclusion}
\label{sec:conclusion}

In this paper, we propose to use hyper-representations as pre-training for transfer learning.
We extend the training objective of hyper-representations by a novel layer-wise loss normalization which is key to the capability of generating functional models.
Our method allows us to generate populations of model weights in a single forward pass.
We evaluate sampled models both in-dataset as well as in transfer learning and find them capable to outperform both models trained from scratch, as well as pre-trained and fine-tuned models.
Our work might serve as a building block for transfer learning from different domains, meta learning or continual learning.\looseness-1

\bibliographystyle{plainnat}
\bibliography{./bibliography.bib}

\begin{thebibliography}{24}
\providecommand{\natexlab}[1]{#1}
\providecommand{\url}[1]{\texttt{#1}}
\expandafter\ifx\csname urlstyle\endcsname\relax
  \providecommand{\doi}[1]{doi: #1}\else
  \providecommand{\doi}{doi: \begingroup \urlstyle{rm}\Url}\fi

\bibitem[Chen et~al.(2019)Chen, Liu, Kira, Wang, and Huang]{chen2019closer}
Wei-Yu Chen, Yen-Cheng Liu, Zsolt Kira, Yu-Chiang~Frank Wang, and Jia-Bin
  Huang.
\newblock A closer look at few-shot classification.
\newblock \emph{arXiv preprint arXiv:1904.04232}, 2019.

\bibitem[Coates et~al.(2011)Coates, Lee, and
  Ng]{coatesAnalysisSingleLayerNetworks2011}
Adam Coates, Honglak Lee, and Andrew~Y Ng.
\newblock An {{Analysis}} of {{Single-Layer Networks}} in {{Unsupervised
  Feature Learning}}.
\newblock In \emph{Proceedings of the 14th {{International Con-}} Ference on
  {{Artificial Intelligence}} and {{Statistics}} ({{AISTATS}})}, page~9, 2011.

\bibitem[Deutsch(2018)]{deutschGeneratingNeuralNetworks2018}
Lior Deutsch.
\newblock Generating {{Neural Networks}} with {{Neural Networks}}.
\newblock \emph{arXiv:1801.01952 [cs, stat]}, April 2018.

\bibitem[Dhillon et~al.(2019)Dhillon, Chaudhari, Ravichandran, and
  Soatto]{dhillon2019baseline}
Guneet~S Dhillon, Pratik Chaudhari, Avinash Ravichandran, and Stefano Soatto.
\newblock A baseline for few-shot image classification.
\newblock \emph{arXiv preprint arXiv:1909.02729}, 2019.

\bibitem[Ghosh et~al.(2020)Ghosh, Sajjadi, Vergari, Black, and
  Sch{\"o}lkopf]{ghoshVariationalDeterministicAutoencoders2020}
Partha Ghosh, Mehdi S.~M. Sajjadi, Antonio Vergari, Michael Black, and Bernhard
  Sch{\"o}lkopf.
\newblock From {{Variational}} to {{Deterministic Autoencoders}}.
\newblock \emph{arXiv:1903.12436 [cs, stat]}, May 2020.

\bibitem[Glorot and Bengio(2010)]{glorotUnderstandingDifficultyTraining2010}
Xavier Glorot and Yoshua Bengio.
\newblock Understanding the difficulty of training deep feedforward neural
  networks.
\newblock page~8, 2010.

\bibitem[Guo et~al.(2019)Guo, Chen, Chen, Wu, Shi, and Tan]{guo2019auto}
Yong Guo, Qi~Chen, Jian Chen, Qingyao Wu, Qinfeng Shi, and Mingkui Tan.
\newblock Auto-embedding generative adversarial networks for high resolution
  image synthesis.
\newblock \emph{IEEE Transactions on Multimedia}, 21\penalty0 (11):\penalty0
  2726--2737, 2019.

\bibitem[Ha et~al.(2016)Ha, Dai, and Le]{haHyperNetworks2016}
David Ha, Andrew Dai, and Quoc~V. Le.
\newblock {{HyperNetworks}}.
\newblock \emph{arXiv:1609.09106 [cs]}, December 2016.

\bibitem[He et~al.(2015)He, Zhang, Ren, and Sun]{heDelvingDeepRectifiers2015}
Kaiming He, Xiangyu Zhang, Shaoqing Ren, and Jian Sun.
\newblock Delving {{Deep}} into {{Rectifiers}}: {{Surpassing Human-Level
  Performance}} on {{ImageNet Classification}}.
\newblock \emph{arXiv:1502.01852 [cs]}, February 2015.

\bibitem[He et~al.(2019)He, Girshick, and Doll{\'a}r]{he2019rethinking}
Kaiming He, Ross Girshick, and Piotr Doll{\'a}r.
\newblock Rethinking imagenet pre-training.
\newblock In \emph{Proceedings of the IEEE/CVF International Conference on
  Computer Vision}, pages 4918--4927, 2019.

\bibitem[Knyazev et~al.(2021)Knyazev, Drozdzal, Taylor, and
  {Romero-Soriano}]{knyazevParameterPredictionUnseen2021}
Boris Knyazev, Michal Drozdzal, Graham~W. Taylor, and Adriana {Romero-Soriano}.
\newblock Parameter {{Prediction}} for {{Unseen Deep Architectures}}.
\newblock \emph{arXiv:2110.13100 [cs, stat]}, October 2021.

\bibitem[Kolesnikov et~al.(2020)Kolesnikov, Beyer, Zhai, Puigcerver, Yung,
  Gelly, and Houlsby]{kolesnikov2020big}
Alexander Kolesnikov, Lucas Beyer, Xiaohua Zhai, Joan Puigcerver, Jessica Yung,
  Sylvain Gelly, and Neil Houlsby.
\newblock Big transfer (bit): General visual representation learning.
\newblock In \emph{European conference on computer vision}, pages 491--507.
  Springer, 2020.

\bibitem[Krizhevsky(2009)]{krizhevskyLearningMultipleLayers2009}
Alex Krizhevsky.
\newblock Learning {{Multiple Layers}} of {{Features}} from {{Tiny Images}}.
\newblock page~60, 2009.

\bibitem[Lecun et~al.(1998)Lecun, Bottou, Bengio, and
  Haffner]{lecunGradientbasedLearningApplied1998}
Y.~Lecun, L.~Bottou, Y.~Bengio, and P.~Haffner.
\newblock Gradient-based learning applied to document recognition.
\newblock \emph{Proceedings of the IEEE}, 86\penalty0 (11):\penalty0
  2278--2324, November 1998.
\newblock ISSN 1558-2256.
\newblock \doi{10.1109/5.726791}.

\bibitem[Liu et~al.(2019)Liu, Yao, and Ren]{liu2019acceleration}
Jinlin Liu, Yuan Yao, and Jianqiang Ren.
\newblock An acceleration framework for high resolution image synthesis.
\newblock \emph{arXiv preprint arXiv:1909.03611}, 2019.

\bibitem[Martin et~al.(2021)Martin, Peng, and Mahoney]{martin2021predicting}
Charles~H Martin, Tongsu~Serena Peng, and Michael~W Mahoney.
\newblock Predicting trends in the quality of state-of-the-art neural networks
  without access to training or testing data.
\newblock \emph{Nature Communications}, 12\penalty0 (1):\penalty0 1--13, 2021.

\bibitem[Mishkin and Matas(2015)]{mishkin2015all}
Dmytro Mishkin and Jiri Matas.
\newblock All you need is a good init.
\newblock \emph{arXiv preprint arXiv:1511.06422}, 2015.

\bibitem[Netzer et~al.(2011)Netzer, Wang, Coates, Bissacco, Wu, and
  Ng]{netzerReadingDigitsNatural2011}
Yuval Netzer, Tao Wang, Adam Coates, Alessandro Bissacco, Bo~Wu, and Andrew~Y
  Ng.
\newblock Reading {{Digits}} in {{Natural Images}} with {{Unsupervised Feature
  Learning}}.
\newblock In \emph{{{NIPS Workshop}} on {{Deep Learning}} and {{Unsupervised
  Feature Learning}} 2011}, page~9, 2011.

\bibitem[Rasmus et~al.(2015)Rasmus, Berglund, Honkala, Valpola, and
  Raiko]{rasmus2015semi}
Antti Rasmus, Mathias Berglund, Mikko Honkala, Harri Valpola, and Tapani Raiko.
\newblock Semi-supervised learning with ladder networks.
\newblock \emph{Advances in neural information processing systems}, 28, 2015.

\bibitem[Ratzlaff and Fuxin(2019)]{ratzlaffHyperGANGenerativeModel2019}
Neale Ratzlaff and Li~Fuxin.
\newblock {{HyperGAN}}: {{A Generative Model}} for {{Diverse}}, {{Performant
  Neural Networks}}.
\newblock In \emph{Proceedings of the 36th {{International Conference}} on
  {{Machine Learning}}}, pages 5361--5369. {PMLR}, May 2019.

\bibitem[Sch{\"u}rholt et~al.(2021)Sch{\"u}rholt, Kostadinov, and
  Borth]{schurholtSelfSupervisedRepresentationLearning2021}
Konstantin Sch{\"u}rholt, Dimche Kostadinov, and Damian Borth.
\newblock Self-{{Supervised Representation Learning}} on {{Neural Network
  Weights}} for {{Model Characteristic Prediction}}.
\newblock In \emph{{{NeurIPS}})}, volume~35, page~13, 2021.

\bibitem[Unterthiner et~al.(2020)Unterthiner, Keysers, Gelly, Bousquet, and
  Tolstikhin]{unterthinerPredictingNeuralNetwork2020}
Thomas Unterthiner, Daniel Keysers, Sylvain Gelly, Olivier Bousquet, and Ilya
  Tolstikhin.
\newblock Predicting {{Neural Network Accuracy}} from {{Weights}}.
\newblock \emph{arXiv:2002.11448 [cs, stat]}, February 2020.

\bibitem[Zhang et~al.(2020)Zhang, Ren, and
  Urtasun]{zhangGraphHyperNetworksNeural2020}
Chris Zhang, Mengye Ren, and Raquel Urtasun.
\newblock Graph {{HyperNetworks}} for {{Neural Architecture Search}}.
\newblock \emph{arXiv:1810.05749 [cs, stat]}, December 2020.

\bibitem[Zhmoginov et~al.(2022)Zhmoginov, Sandler, and
  Vladymyrov]{zhmoginovHyperTransformerModelGeneration2022}
Andrey Zhmoginov, Mark Sandler, and Max Vladymyrov.
\newblock {{HyperTransformer}}: {{Model Generation}} for {{Supervised}} and
  {{Semi-Supervised Few-Shot Learning}}.
\newblock \emph{arXiv:2201.04182 [cs]}, January 2022.

\end{thebibliography}

\newpage
\appendix
\onecolumn

\section{Model Zoo Details}
\label{app:zoos}
All model zoos share one general CNN architecture, outlined in Table \ref{tab:model_zoo_architecture}. 
The hyperparameter choices for each of the population are listed in Table \ref{tab:model_zoo_hyperparameters}. 
The hyperparameters are chosen to generate zoos with smooth, continuous development and spread in performance. 
The models in the zoos trained on MNIST, SVHN and USPS have one input channel and 2464 learnable parameters. 
Due to the three input channels in the CIFAR-10 and STL-10 zoos, their models have 2864 learnable parameters.

\%
\begin{table}[h!]
    \begin{minipage}{0.48\textwidth}
    \centering
    {\small
    \begin{tabular}{l|l|l}
        \toprule
        \textbf{Layer}          & \textbf{Component} & \textbf{Value} \\
        \midrule 
        \multirow{5}{*}{Conv 1} & input channels     & 1/3                 \\
                                & output channels    & 8                    \\
                                & kernel size        & 5                    \\
                                & stride             & 1                    \\
                                & padding            & 0                    \\
        \midrule
        Max Pooling             & kernel size        & 2                    \\
        \midrule
        Activation              & tanh / gelu         &                      \\
        \midrule
        \multirow{5}{*}{Conv 2} & input channels     & 8                    \\
                                & output channels    & 6                    \\
                                & kernel size        & 5                    \\
                                & stride             & 1                    \\
                                & padding            & 0                    \\
        \midrule
        Max Pooling             & kernel size        & 2                    \\
        \midrule
        Activation              & tanh / gelu         &                      \\
        \midrule
        \multirow{5}{*}{Conv 3} & input channels     & 6                    \\
                                & output channels    & 4                    \\
                                & kernel size        & 2                    \\
                                & stride             & 1                    \\
                                & padding            & 0                    \\
        \midrule
        Activation              & tanh / gelu         &                      \\
        \midrule
        \multirow{2}{*}{Linear 1} & input channels     & 36                   \\
                                & output channels    & 20                   \\
        \midrule
        Activation              & tanh / gelu         &                      \\
        \midrule
        \multirow{2}{*}{Linear 2} & input channels     & 20                   \\
                                & output channels    & 10                   \\
        \bottomrule
    \end{tabular}
    \label{tab:model_zoo_architecture}
    \caption{CNN architecture details for the models in model zoos. }
    }
    \end{minipage}
    \hspace{2mm}
    \begin{minipage}{0.48\textwidth}
    \centering
    {\small
    \begin{tabular}{l|l|l}
        \toprule
        \textbf{Model Zoo}     & \textbf{Hyperparameter} & \textbf{Value}   \\
        \midrule
        \multirow{5}{*}{MNIST} & input channels          & 1                \\
                               & activation              & tanh             \\
                               & weight decay                      & 0                \\
                               & learning rate                      & 3e-4             \\
                               & initialization          & uniform          \\
                               & optimizer               & Adam             \\
                               & seed                    & [1-1000]         \\
        \midrule
        \multirow{5}{*}{SVHN}  & input channels          & 1                \\
                               & activation              & tanh             \\
                               & weight decay                      & 0                \\
                               & learning rate                      & 3e-3             \\
                               & initialization          & uniform          \\
                               & optimizer               & adam             \\
                               & seed                    & [1-1000]         \\
        \midrule
        \multirow{5}{*}{USPS}  & input channels          & 1                \\
                               & activation              & tanh             \\
                               & weight decay                      & 1e-3             \\
                               & learning rate                      & 1e-4             \\
                               & initialization          & kaiming\_uniform \\
                               & optimizer               & adam             \\
                               & seed                    & [1-1000]         \\
        \midrule
        \multirow{5}{*}{CIFAR-10} & input channels          & 3                \\
                               & activation              & gelu             \\
                               & weight decay                      & 1e-2             \\
                               & learning rate                      & 1e-4             \\
                               & initialization          & kaiming-uniform  \\
                               & optimizer               & adam             \\
                               & seed                    & [1-1000]         \\
        \midrule
        \multirow{5}{*}{STL-10}& input channels          & 3                \\
                               & activation              & tanh             \\
                               & weight decay                      & 1e-3             \\
                               & learning rate                      & 1e-4             \\
                               & initialization          & kaiming-uniform  \\
                               & optimizer               & adam             \\
                               & seed                    & [1-1000]         \\
        \bottomrule
    \end{tabular}
    \label{tab:model_zoo_hyperparameters}
    \caption{Hyperparameter choices for the model zoos. }
    
    }
    \end{minipage}
\end{table}
\newpage
\section{Cross Dataset Initialization Results }
\label{app:results}
In Table \ref{tab:transfer_2} we report the remaining results on cross dataset initialization. On MNIST to SVHN, the sampled population outperforms both baselines. On CIFAR-10 to STL-10, much like the other way around, all populations saturate.
\begin{table}[ht!]
\centering
\begin{minipage}[t]{0.45\textwidth}
\centering
\caption{Transfer-learning results (mean and std accuracy in \%). Note that for CIFAR-10 to STL-10 the performance of all methods saturate quickly due to the limited capacity of models in the zoo making further improvements challenging.
}
\label{tab:transfer_2}
\vspace{2mm}
\scriptsize
\setlength{\tabcolsep}{10pt}
\begin{tabularx}{\linewidth}{rccc}
\toprule
\multicolumn{1}{r}{\textbf{Method}} & \textbf{Ep.} & \multicolumn{1}{c}{\textbf{MNIST to SVHN}} & \multicolumn{1}{c}{\textbf{CIFAR-10 to STL-10}} \\
\midrule
$B_{T}$                & 0      &  \multicolumn{2}{c}{$\approx$10\% (random guessing)}\\
$B_{F}$                & 0      & 14.9 \textit{(2.8)}    & \textbf{25.2 \textit{(1.1)}}          \\
$S_{\text{KDE}30}$     & 0      & \textbf{15.9 \textit{(2.7)}} & 13.7 \textit{(2.0)}             \\
\midrule
$B_{T}$                & 1      & 19.4 \textit{(0.6)} & 15.4 \textit{(1.8)}   \\
$B_{F}$                & 1      & \textbf{21.9 \textit{(4.1)}} & \textbf{26.0 \textit{(1.1)}}   \\
$S_{\text{KDE}30}$     & 1      & 20.8 \textit{(2.7)} & 19.2 \textit{(1.2)}   \\
\midrule
$B_{T}$                & 50     & 70.7 \textit{(8.8)} & 39.0 \textit{(1.0)}  \\
$B_{F}$                & 50     & 76.1 \textit{(1.4)} & \textbf{42.7 \textit{(1.2)}}  \\
$S_{\text{KDE}30}$     & 50     & \textbf{77.1 \textit{(1.5)}} & 41.1 \textit{(0.9)}  \\
\bottomrule 
\end{tabularx}
\end{minipage}
\end{table} 
\paragraph{Limitations of Zoos with Small Models}
To thoroughly investigate different methods and make experiments feasible, we chose to use the model zoos of the same scale as in~\citep{schurholtSelfSupervisedRepresentationLearning2021}. 
While on MNIST and SVHN, the architectures of such model zoos allowed us to achieve a reasonably high performance, on CIFAR-10 and STL-10, the performance of all populations is limited by the low capacity of the models architecture. The models saturate at around 50\% and 40\% accuracy, respectively. 
We hypothesize that due to the high remaining loss, the weight updates are correspondingly large without converging or improving performance. 
This may cause the weights to contain relatively little signal and high noise. Indeed, learning hyper-representations on the CIFAR-10 and STL-10 zoos was difficult and never reached similar reconstruction performance as in the MNIST and SVHN zoos. That in turn additionally limits the performance of the sampled populations.
Future work will therefore focus on architectures with higher capacity.

\end{document}